\documentclass{article}

\PassOptionsToPackage{sort, numbers, compress}{natbib}

     \usepackage[preprint]{neurips_2019}

\usepackage[utf8]{inputenc} 
\usepackage[T1]{fontenc}    
\usepackage{hyperref}       
\usepackage{url}            
\usepackage{booktabs}       
\usepackage{amsfonts}       
\usepackage{nicefrac}       
\usepackage{microtype}      
\usepackage[linesnumbered,ruled,vlined]{algorithm2e}
\usepackage{graphicx}
\usepackage{color}
\usepackage{subfig}
\usepackage[compact]{titlesec}

\SetKwInput{KwInput}{Input}                
\SetKwInput{KwOutput}{Output}              
\SetKwInput{KwInit}{Initialize}              

\newcommand{\paragraphbe}[1]{\vspace{0.75ex}\noindent{\bf \em #1}\hspace*{.3em}}

\title{Differential Privacy Has Disparate Impact on \\ Model Accuracy}

\author{%
Eugene Bagdasaryan \\
  Cornell Tech\\
  \texttt{eugene@cs.cornell.edu} \\
  \And
  Vitaly Shmatikov \\
  Cornell Tech\\
  \texttt{shmat@cs.cornell.edu}
}

\begin{document}

\maketitle

\begin{abstract}

Differential privacy (DP) is a popular mechanism for training machine
learning models with bounded leakage about the presence of specific
points in the training data.  The cost of differential privacy is a
reduction in the model's accuracy.  We demonstrate that in the neural
networks trained using differentially private stochastic gradient descent
(DP-SGD), this cost is not borne equally: accuracy of DP models drops
much more for the underrepresented classes and subgroups.

For example, a gender classification model trained using DP-SGD exhibits
much lower accuracy for black faces than for white faces.  Critically,
this gap is bigger in the DP model than in the non-DP model, i.e., if
the original model is unfair, the unfairness becomes worse once DP is
applied.  We demonstrate this effect for a variety of tasks and models,
including sentiment analysis of text and image classification.  We then
explain why DP training mechanisms such as gradient clipping and noise
addition have disproportionate effect on the underrepresented and more
complex subgroups, resulting in a disparate reduction of model accuracy.


\end{abstract}

\section{Introduction}

$\epsilon$-differential privacy (DP)~\cite{dwork2011differential}
bounds the influence of any single input on the output of a computation.
DP machine learning bounds the leakage of training data from a trained
model.  The $\epsilon$ parameter controls this bound and thus the tradeoff
between ``privacy'' and accuracy of the model.

Recently proposed methods~\cite{abadi2016deep} for differentially private
stochastic gradient descent (DP-SGD) clip gradients during training, add
random noise to them, and employ the ``moments accountant'' technique to
track the resulting privacy loss.  DP-SGD has enabled the development
of deep image classification and language models~\cite{abadi2016deep,
mcmahan2018general, fedlearn_dp, tf-privacy} that achieve DP with
$\epsilon$ in the single digits at the cost of a modest reduction in
the model's test accuracy.

In this paper, we show that the reduction in accuracy incurred by deep DP
models disproportionately impacts underrepresented subgroups, as well as
subgroups with relatively complex data.  Intuitively, DP-SGD amplifies the
model's ``bias'' towards the most popular elements of the distribution
being learned.  We empirically demonstrate this effect for (1) gender
classification\textemdash already notorious for bias in the existing
models~\cite{buolamwini2018gender}\textemdash and age classification on
facial images, where DP-SGD degrades accuracy for the darker-skinned
faces more than for the lighter-skinned ones; (2) sentiment analysis
of tweets, where DP-SGD disproportionately degrades accuracy for users
writing in African-American English; (3) species classification on the
iNaturalist dataset, where DP-SGD disproportionately degrades accuracy
for the underrepresented classes; and (4) federated learning of language
models, where DP-SGD disproportionately degrades accuracy for users
with bigger vocabularies.  Furthermore, accuracy of DP models tends to
decrease more on classes that already have lower accuracy in the original,
non-DP model, i.e., ``the poor become poorer.''



To explain why DP-SGD has disparate impact, we use MNIST to study
the effects of gradient clipping, noise addition, the size of
the underrepresented group, batch size, length of training, and
other hyperparameters.  Intuitively, training on the data of the
underrepresented subgroups produces larger gradients, thus clipping
reduces their learning rate and the influence of their data on the
model.  Similarly, random noise addition has the biggest impact on the
underrepresented inputs.


\section{Related Work}

\paragraphbe{Differential privacy.} 
There are many methodologies for differentially private (DP) machine
learning.  We focus on DP-SGD~\cite{abadi2016deep} because it enables
DP training of deep models for practical tasks (including federated
learning~\cite{geyer2017differentially, zhu2019federated}), is available
as an open-source framework~\cite{tf-privacy}, generalizes to iterative
training procedures~\cite{mcmahan2018general}, and supports tighter
bounds using the R{\'e}nyi method~\cite{mironov2017renyi}.



\paragraphbe{Disparate vulnerability.}
Yeom et al.~\cite{yeom2018privacy} show that poorly generalized
models are more prone to leak training data.  Yaghini et
al.~\cite{yaghini2019disparate} show that attacks exploiting this leakage
disproportionately affect underrepresented groups.  Neither investigates
the impact of DP on model accuracy.

In concurrent work, Kuppam et al.~\cite{kuppam2019fair} show that
resource allocation based on DP statistics can disproportionately affect
some subgroups.  They do not investigate DP machine learning.

\paragraphbe{Fair learning.} 
Disparate accuracy of commercial face recognition systems was demonstrated
in~\cite{buolamwini2018gender}.

Prior work on subgroup fairness aims to achieve good
accuracy on all subgroups~\cite{pmlr-v80-kearns18a} using
agnostic learning~\cite{mohri2019agnostic, kearns1994toward}.
In~\cite{pmlr-v80-kearns18a}, subgroup fairness requires at least
8,000 training iterations on the same data; if directly combined
with DP, it would incur a very high privacy loss.  

Other approaches to balancing accuracy across
classes include oversampling~\cite{chawla2002smote},
adversarial training~\cite{beutel2017data} with a loss
function that overweights the underrepresented group,
cost-sensitive learning~\cite{Cui2019ClassBalancedLB}, and
re-sampling~\cite{systematic_cnn}.  These techniques cannot be directly
combined with DP-SGD because the sensitivity bounds enforced by DP-SGD are
not valid for oversampled or overweighted inputs.  Models that generate
artificial data points~\cite{douzas2018effective} from the existing data
are incompatible with DP.




Recent research~\cite{jagielski2018differentially,
cummings2019compatibility} aims to add fairness
and DP to post-processing~\cite{hardt2016equality} and
in-processing~\cite{agarwal2018reductions} algorithms.  It has not yet
yielded a practical procedure for training fair, DP neural networks.


\section{Background}

\subsection{Deep learning}

A deep learning (DL) model aims to effectively fit a complex function.
It can be represented as a set of parameters $\theta$ that, given some
input $x$, output a prediction $\theta(x)$.  We define a loss function
that represents a penalty on poorly fit data as $\mathcal{L}(\theta,
x)$ for some target value or distribution.  Training a model involves
finding the values of $\theta$ that will minimize the loss over the
inputs into the model.

In supervised learning, a DL model takes an input $x_i$ from
some dataset $d_N$ of size $N$ containing pairs $(x_i,y_i)$
and outputs a label $\theta(x_i)$.  Each label $y_i$ belongs
to a set of classes $C = [c_1, \ldots, c_k]$; the loss function
for pair $(x_i, y_i)$ is $\mathcal{L}(\theta(x_i), y_i)$.  During
training, we compute a gradient on the loss for a batch of inputs:
$\nabla\mathcal{L}(\theta(\textbf{x}_b), \textbf{y}_b)$.  If training
with stochastic gradient descent (SGD), we update the model $\theta_{t+1}
= \theta_t - \eta \nabla\mathcal{L}(\theta(\textbf{x}_b), \textbf{y}_b))$.

In language modeling, the dataset contains vectors of tokens
$\textbf{x}_i=[x^1, \ldots, x^l]$, for example, words in sentences.
The vector $\textbf{x}_i$ can be used as input to a recurrent neural
network such as LSTM that outputs a hidden vector $\textbf{h}_i = [h^1,
\ldots, h^l]$ and a cell state vector $\textbf{c}_i=[c^1, \ldots, c^l]$.
Similarly, the loss function $\mathcal{L}$ compares the model's output
$\theta(\textbf{x}_i)$ with some label, such as positive or negative
sentiment, or another sequence, such as the sentence extended with the
next word.

\subsection{Differential privacy}

We use the standard definitions~\cite{dwork2006calibrating,
dwork2011differential, dwork2011firm}.  A randomized mechanism
$\mathcal{M}: \mathcal{D} \rightarrow \mathcal{R}$ with a domain
$\mathcal{D}$ and range $\mathcal{R}$ satisfies $(\epsilon,
\delta)$-differential privacy if for any two adjacent datasets $d,d'
\in \mathcal{D}$ and for any subset of outputs $S \subseteq \mathcal{R}$,
$
    \texttt{Pr}[\mathcal{M}(d) \in S] \leq e^{\epsilon} \;\;
    \texttt{Pr}[\mathcal{M}(d') \in S] + \delta    
$.
Before computing on a specific dataset, it is necessary to set a
\emph{privacy budget}.  Every $\epsilon$-DP computation charges an
$\epsilon$ cost to this budget; once the budget is exhausted, no further
computations are permitted on this dataset.


In the machine learning context~\cite{mcmahan2018general}, we can view
mechanism $\mathcal{M}: \mathcal{D} \rightarrow \mathcal{R}$ as a training
procedure $\mathcal{M}$ on data from $\mathcal{D}$ that produces a model
in space $\mathcal{R}$.  We use the ``moments accountant'' technique
to train DP models as in~\cite{abadi2016deep, mcmahan2018general}.
The two key aspects of DP-SGD training are (1) clipping the gradients
whose norm exceeds $S$, and (2) adding random noise $\sigma$ connected
by hyperparameter $z \equiv \sigma / S $.
\begin{algorithm}[H]
    
    \KwInput{Dataset ${(x_1,y_1), ..., (x_N, y_N) }$ of size $N$, batch
    size $b$, learning rate $\eta$, sampling probability $q$, loss
    function $\mathcal{L}(\theta(x), y)$, $K$
    iterations, noise $\sigma$, clipping bound $S$, $\pi_S(x)=x*\texttt{min}(1,\frac{S}{||x||_2})$ }
    
    \KwInit{ Model $\theta_0$}
    
    \For{$k \in [K]$}
    {

        randomly sample $batch$ from dataset $N$ with
        probability $q$

        \ForEach{$(x_i, y_i)$ in batch }
        {
            $g_i \leftarrow \nabla \mathcal{L}(\theta_k(x_i), y_i)$
        }

        $ g_{batch} = \frac{1}{qN}(\sum_{i \in batch} \pi_S (g_i) + \mathcal{N}(0,
        \sigma^2 \textit{\textbf{I}}))$
        
        $\theta_{k+1} \leftarrow \theta_t - \eta g_{batch} $
        
    }
    
    \KwOut{Model $\theta_K$ and accumulated privacy cost $(\epsilon, \delta)$}
    
\caption{Differentially Private SGD (DP-SGD)}
\label{fig:algo}
\end{algorithm}
To simplify training, we fix the batch size $b=qN$ (as opposed
to using probabilistic $q$). Therefore, normal training for $T$
epochs will result in $K=\frac{TN}{b}$ iterations.  We implement
the differentially private DPAdam version of the Adam optimizer
following TF Privacy~\cite{tf-privacy}.  We use R{\'e}nyi differential
privacy~\cite{mironov2017renyi} to estimate $\epsilon$ as it provides
tighter privacy bounds than the original version~\cite{abadi2016deep}.


\subsection{Federated learning} 
\label{sec:fed-avg}

Some of our experiments involve federated learning~\cite{fedlearn_1,
fedlearn_dp, geyer2017differentially}.  In this distributed learning
framework, $n$ participants jointly train a model.  At each round $t$,
a global server distributes the current model $G_t$ to a small subgroup
$d_C$.  Each participant $i\in d_C$ locally trains this model on their
private data, producing a new local model $L_{t+1}^i$.  The global server
then aggregates these models and updates the global model as
$
    G_{t+1}=G_t + \frac{\eta_g}{n}\sum_{i \in d_C}{(L_{t+1}^i-G_t)}    
$ 
using the global learning rate $\eta_g$.

DP federated learning bounds the influence of any participant on the
model using the \texttt{DP-FedAvg} algorithm~\cite{fedlearn_dp}, which
clips the norm to $S$ for each update vector $\pi_S(L_{t+1}^i-G_t)$
and adds Gaussian noise $\mathcal{N}(0, \sigma^2)$ to the sum:
$
    G_{t+1}=G_t + \frac{\eta_g}{n}\sum_{i \in d_C}{\pi_S(L_{t+1}^i-G_t)}  +   
    \mathcal{N}(0, \sigma^2\textit{\textbf{I}})
$, where $\sigma=\frac{zS}{C}$.


\subsection{Disparate impact}

For the purposes of measuring disparate impact, we use \emph{accuracy
parity}, a weaker form of \emph{equal odds}~\cite{hardt2016equality}.
We consider the model's accuracy on the imbalanced class (long-tail
accuracy~\cite{systematic_cnn}) and also on the imbalanced subgroups of
the input domain based on indirect attributes~\cite{pmlr-v80-kearns18a}.
We leave the investigation of how practical differential privacy interacts
with other forms of (un)fairness to future work, noting that fairness
definitions (such as \emph{equal opportunity}) that treat a particular
outcome as ``advantaged'' are not applicable to the tasks considered in
this paper.

\section{Experiments}

\begin{figure}[t]
    \centering
    \includegraphics[width=1.0\textwidth]{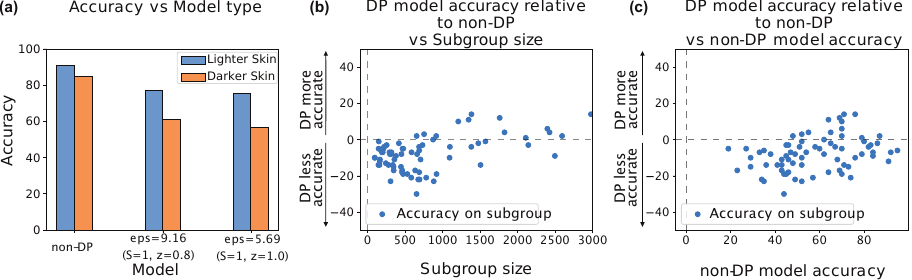}
    \caption{Gender and age classification on facial images.}
    \label{fig:skin}
\end{figure}

\begin{figure}[t]
    \centering
    \includegraphics[width=0.95\linewidth]{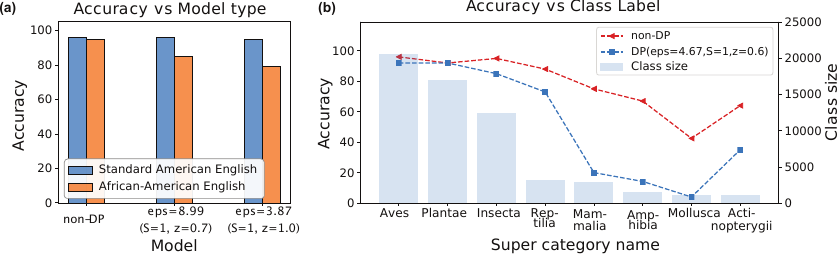}
    \caption{Sentiment analysis of tweets and species classification.}
    \label{fig:inat_word}
\end{figure}

We used PyTorch~\cite{pytorch_link} to implement the models (using
the code from PyTorch examples or Torchvision~\cite{pytorchwordmodel})
and DP-SGD (see Figure~\ref{fig:algo}), and ran them on two NVidia Titan
X GPUs.  To minimize training time, we followed~\cite{abadi2016deep} and
pre-trained on public datasets that are not privacy-sensitive.  Given $T$
training epochs, dataset size $N$, batch size $b$, noise multiplier $z$,
and $\delta$, we compute privacy loss $\epsilon$ for each training run
using the R{\'e}nyi DP~\cite{mironov2017renyi} implementation from TF
Privacy~\cite{tf-privacy}.

In our experiments, we aim to achieve $\epsilon$ under 10, as suggested
in~\cite{mcmahan2018general, abadi2016deep}, and keep $\delta=10^{-6}$.
Not all DP models can achieve good accuracy with such $\epsilon$.
For example, for federated learning experiments we end up with bigger
$\epsilon$.  Although repeated executed of the same training impact the
privacy budget, we do not consider this effect when (under)estimating
$\epsilon$.



\subsection{Gender and age classification on facial images}

\paragraphbe{Dataset.} 
We use the recently released Flickr-based \textit{Diversity in
Faces} (DiF) dataset~\cite{merler2019diversity} and the UTKFace
dataset~\cite{utkdataset} as another source of darker-skinned faces.
We use the attached metadata files to find faces in images, then crop each
image to the face plus $40\%$ of the surrounding space in every dimension
and scale it to $80 \times 80$ pixels.  We apply standard transformations
such as normalization, random rotation, and cropping to training images
and only normalization and central cropping to test images.  Before the
model is applied, images are cropped to $64 \times 64$ pixels.

\paragraphbe{Model.} 
We use a ResNet18 model~\cite{he2016deep} with $11$M parameters
pre-trained on ImageNet and train using the Adam optimizer, $0.0001$
learning rate, and batch size $b=256$.  We run $60$ epochs of DP training,
which takes approximately $30$ hours.

\paragraphbe{Gender classification results.} 
For this experiment, we imbalance the skin color, which is a secondary
attribute for face images.  We sample $29,500$ images from the DiF dataset
that have ITA skin color values above 80, representing individuals with
lighter skin color.  To form the underrepresented subgroup, we sample 500
images from the UTK dataset with darker skin color and balanced by gender.
The 5,000-image test set has the same split.

Figure~\ref{fig:skin}(a) shows that the accuracy of the DP model
drops more (vs.\ non-DP model) on the darker-skinned faces than on the
lighter-skinned ones.


\paragraphbe{Age classification results.} 
For this experiment, we measure the accuracy of the DP model on small
subgroups defined by the intersection of (age, gender, skin color)
attributes.  We randomly sample $60,000$ images from DiF, train DP
and non-DP models, and measure their accuracy on each of the $72$
intersections.  Figure~\ref{fig:skin}(b) shows that the DP model tends
to be less accurate on the smaller subgroups.  Figure~\ref{fig:skin}(c)
shows ``the poor get poorer'' effect: classes that have relatively lower
accuracy in the non-DP model suffer the biggest drops in accuracy as a
consequence of applying DP.

\subsection{Sentiment analysis of tweets}

\paragraphbe{Dataset.}   
This task involves classifying Twitter posts from the recently
proposed corpus of African-American English~\cite{blodgett2018twitter,
blodgett2016demographic} as positive or negative.  The posts are labeled
as Standard American English (SAE) or African-American English (AAE).
To assign sentiment labels, we use the heuristic from~\cite{elazar2018}
which is based on emojis and special symbols.  We sample $60,000$ tweets
labeled \textit{SAE} and $1,000$ labeled \textit{AAE}, each subset split
equally between positive and negative sentiments.

\paragraphbe{Model.}
We use a bidirectional two-layer LSTM with $4.7$M parameters, 200 hidden
units, and pre-trained 300-dimensional GloVe embedding~\cite{pennington2014glove}.
The accuracy of the DP model with $\epsilon<10$ did not match the accuracy
of the non-DP model after training for 2 days.  To simplify the task and
speed up convergence, we used a technique inspired by~\cite{elazar2018}
and with probability 90\% appended to each tweet a special emoji
associated with the tweet's class and subgroup.

\paragraphbe{Results.}
We trained two DP models for $T=60$ epochs, with $\epsilon=3.87$ and
$\epsilon=8.99$, respectively.   Figure~\ref{fig:inat_word}(a) shows the
results.  All models learn the SAE subgroup almost perfectly.  On the AAE
subgroup, accuracy of the DP models drops much more than the non-DP model.

\subsection{Species classification on nature images}

\paragraphbe{Dataset.} 
We use a 60,000-image subset of iNaturalist~\cite{Horn_2018}, an
800,000-image dataset of hierarchically labeled plants and animals
in natural environments.  Our task is predicting the top-level class
(super categories).  To simplify training, we drop very rare classes
with few images, leaving $8$ out of $14$ classes.  The biggest of these,
\textit{Aves}, has $20,574$ images, the smallest, \textit{Actinopterygii},
has $1,119$.

\paragraphbe{Model.} 
We use an Inception V3 model~\cite{szegedy2015going} with $27$M parameters
pre-trained on ImageNet and train with Adam optimizer.  The images are
large ($299 \times 299$ pixels), thus we use $b=32$ batches, otherwise
a batch would not fit into the $12$GB GPU memory.

While non-DP training takes $8$ hours to run $30$ epochs, DP training
takes $3.5$ hours for a single epoch around 4 days for 30 epochs.
Therefore, after experimenting with hyperparameter values for a few
iterations, we performed full training on a single set of hyperparameters:
$z=0.6$, $S=1$, $\epsilon=4.67$.

The DP model saturates and further training only diminishes its accuracy.
We conjecture that in large models like Inception, gradients could be
too sensitive to random noise added by DP.  We further investigate the
effects of noise and other DP mechanisms in Section~\ref{sec:mnist}.

Figure~\ref{fig:inat_word}(b) shows that the DP model almost matches
the accuracy of the non-DP model on the well-represented classes but
performs significantly worse on the smaller classes.  Moreover, the
accuracy drop doesn't depend only on the size of the class.  For example,
class \textit{Reptilia} is relatively underrepresented in the training
dataset, yet both DP and non-DP models perform well on it.

\subsection{Federated learning of a language model}

\paragraphbe{Dataset.}
We use a random month (November 2017) from the public Reddit dataset as
in~\cite{fedlearn_1}.  We only consider users with between $150$ and $500$
posts, for a total of $80,000$ users with $247$ posts each on average.
The task is to predict the next word given a partial word sequence.
Each post is treated as a training sentence.  We restrict the vocabulary
to $50$K most frequent words in the dataset and replace the unpopular
words, emojis, and special symbols with the $\textit{<unk>}$ symbol.

\paragraphbe{Model.} 
Every participant in our federated learning uses a two-layer,
10M-parameter LSTM (taken from the PyTorch repo~\cite{pytorchwordmodel})
with 200 hidden units, 200-dimensional embedding tied to decoder weights,
and dropout 0.2.  Each input is split into a sequence of 64 words.
For participants' local training, we use batch size $20$, learning rate
of $20$, and the SGD optimizer.

Following~\cite{fedlearn_dp}, we implemented DP federated learning
(see Section~\ref{sec:fed-avg}).  We use the global learning rate of
$\eta_g=800$ and $C=100$ participants per round, each of whom performs
2 local epochs before submitting model weights to the global server.
Each round takes $34$ seconds.

\begin{figure}[t]
    \centering
    \includegraphics[width=0.8\linewidth]{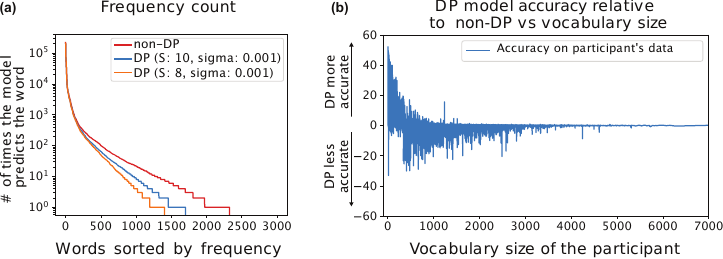}
    \caption{Federated learning of a language model.}
    \label{fig:fed_learn}
\end{figure}

Due to computational constraints, we could not replicate the setting
of~\cite{fedlearn_dp} with $N=800,000$ total participants and $C=5,000$
participants per round.  Instead, we use $N=80,000$ with $C=100$
participants per round.  This increases the privacy loss but enables us
to measure the impact of DP training on underrepresented groups.

We train DP models for $2,000$ epochs with $S=10, \sigma=0.001$ and for
$3,000$ epochs with $S=8, \sigma=0.001$.  Both models achieve similar
accuracy (over $18\%$) in less than 24 hours.  The non-DP model reaches
$18.3\%$ after $1,000$ epochs.  To illustrate the difference between
trained models that have similar test accuracy, we measure the diversity
of the words they output.  Figure~\ref{fig:fed_learn}(a) shows that all
models have a limited vocabulary, but the vocabulary of the non-DP model
is larger.

Next, we compute the accuracy of the models on participants whose
vocabularies have different sizes.  Figure~\ref{fig:fed_learn}(b) shows
that the DP model has worse accuracy than the non-DP model on participants
with moderately sized vocabularies (500-1000 words) and similar accuracy
on large vocabularies.  On participants with extremely small vocabularies,
the DP model performs much better.  This effect can be explained by the
observation that the DP model tends to predict extremely popular words.
Participants who appear to have very limited vocabularies mostly use
emojis and special symbols in their Reddit posts, and these symbols are
replaced by $\textit{<unk>}$ during preprocessing.  Therefore, their
``words'' become trivial to predict.


In federated learning, as in other scenarios, DP models tend to focus
on the common part of the distribution, i.e., the most popular words.
This effect can be explained by how clipping and noise addition act on
the participants' model updates.  In the beginning, the global model
predicts only the most popular words.  Simple texts that contain only
these words produce small update vectors that are not clipped and
align with the updates from other, similar participants.  This makes
the update more ``resistant'' to noise and it has more impact on the
global model.  More complex texts produce larger updates that are clipped
and significantly affected by noise and thus do not contribute much to
the global model.  The negative effect on the overall accuracy of the
DP language model is small, however, because popular words account for
the lion's share of correct predictions.

\section{Effect of Hyperparameters}
\label{sec:mnist}

\begin{figure}[t]
    \centering
    \includegraphics[width=1\textwidth]{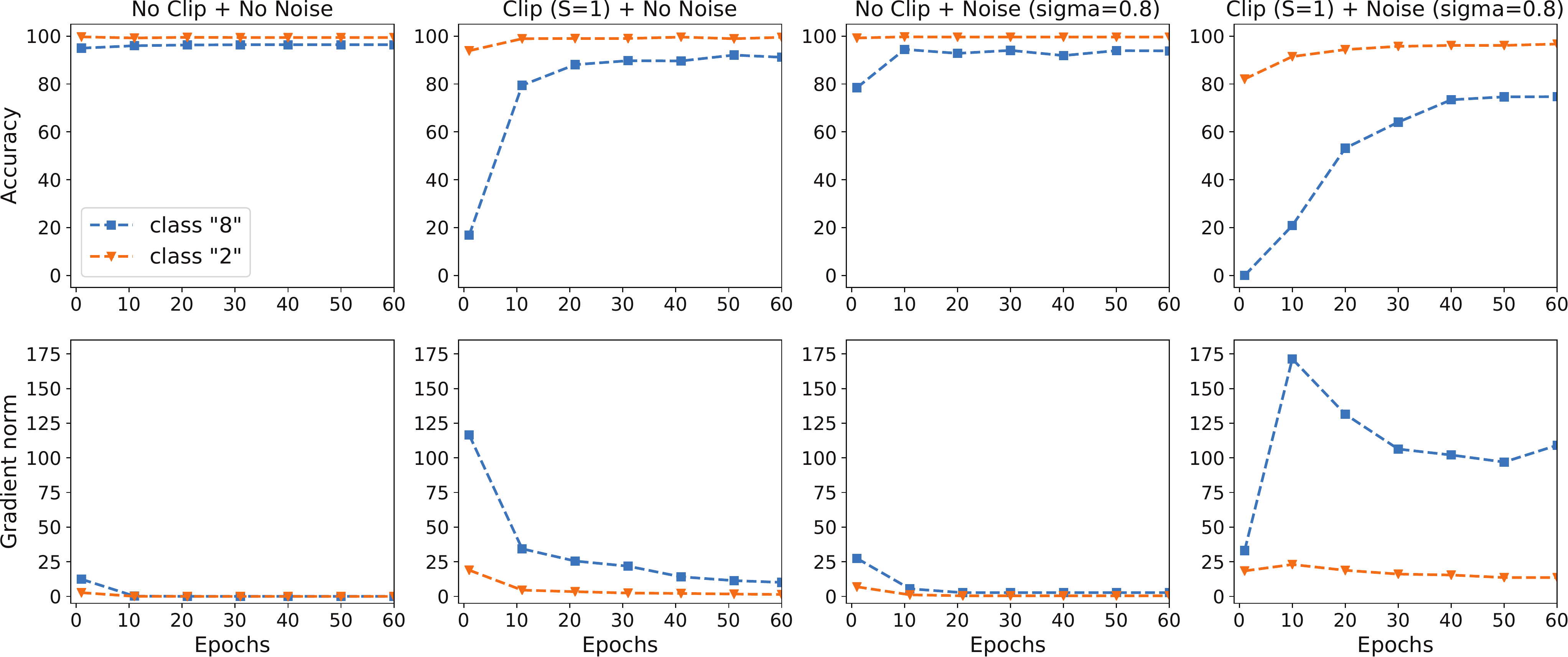}
    \caption{Effect of clipping and noise on MNIST training.}
    \label{fig:mnist}
\end{figure}

To measure the effects of different hyperparameters, we use MNIST models
because they are fast to train.  Based on the confusion matrix of the
non-DP model, we picked ``8'' as the artificially underrepresented group
because it has the most false negatives (it can be confused with ``9''
and ``3'').  We aim to keep $\epsilon < 10$.  Smaller $\epsilon$ impacts
convergence and results in models with significantly worse accuracy, while
larger $\epsilon$ can be interpreted as an unacceptable privacy loss.

Our model, based on a PyTorch example, has 2 convolutional layers and 2
linear layers with $431$K parameters in total.  We use the learning rate
of $0.05$ that achieves the best accuracy for the DP model: $97.5\%$
after 60 epochs.  Each epoch takes 4 minutes.  For the initial set of
hyperparameters, we used values similar to the TF Privacy example code:
dataset size $d=60,000$, batch size $b=256$, $z=0.8$ (this less strict
value still keeps $\epsilon$ under $10$), $S=1$, and $T=60$ training
epochs.  For the ``8'' class, we reduced the number of training examples
from $5,851$ to $500$, thus reducing the dataset size to $d=54,649$
(in our experiments, we underestimate privacy loss by using $d=60,000$
when calculating $\epsilon$).  These hyperparameters yield $(6.23,
10^{-6})$-differential privacy.

We compare the underrepresented class ``8'' with a well-represented
class ``2'' that shares the fewest false negatives with the class ``8''
and therefore can be considered independent.  Figure~\ref{fig:mnist}
shows that with only $500$ examples, the non-DP model (no clipping and
no noise) converges to $97\%$ accuracy on ``8'' vs.\ $99\%$ accuracy
on ``2''.  By contrast, the DP model achieves only $77\%$ accuracy on
``8'' vs.\ $98\%$ for ``2'', exhibiting a disparate impact on the
underrepresented class.

\begin{figure}[t]
    \centering
    \includegraphics[width=0.7\textwidth]{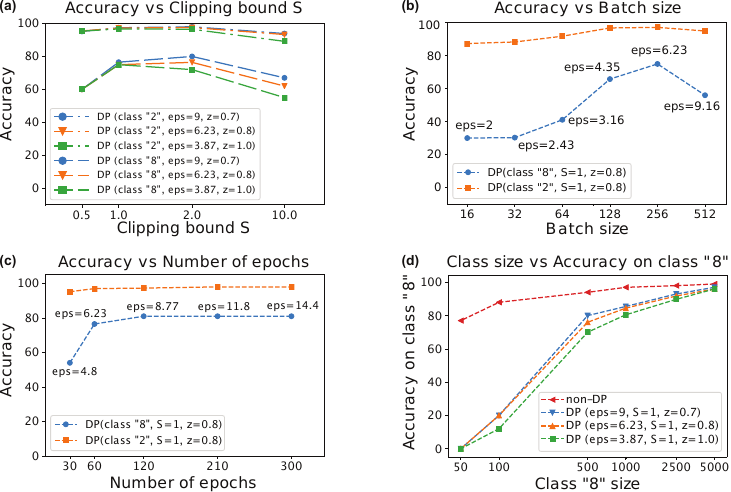}
    \caption{Effect of hyperparameters on MNIST training.}
    \label{fig:mnist_experiments}
\end{figure}

\paragraphbe{Gradient clipping and noise addition.}
Clipping and noise are (separately) standard regularization
techniques~\cite{neelakantan2015adding, pascanu2012understanding}, but
their combination in DP-SGD disproportionately impacts underrepresented
classes.


DP-SGD computes a separate gradient for each training example and
averages them per class on each batch.  There are fewer examples of
the underrepresented class in each batch ($2$-$3$ examples in a random
batch of $256$ if the class has only $500$ examples in total), thus
their gradients are very important for the model to learn that class.



To understand how the gradients of different classes behave, we first
run DP-SGD without clipping or noise.  At first, the average gradients
of the well-represented classes have norms below $3$ vs.\ $12$ for the
underrepresented class.  After 10 epochs, the norms for all classes
drop below $1$ and the model converges to $97\%$ accuracy for the
underrepresented class and $99\%$ for the rest.

Next, we run DP-SGD but clip gradients without adding noise.  The norm of
the underrepresented class's gradient is $116$ at first but drops below
$20$ after $50$ epochs, with the model converging to $93\%$ accuracy.
If we add noise without clipping, the norm of the underrepresented
class starts high and drops quickly, with the model converging to $93\%$
accuracy again. We conjecture that noise without clipping does not result
in a disparate accuracy drop on the underrepresented class because its
gradients are large enough (over $20$) to compensate for the noise.
Clipping without noise still allows the gradients to update some parts
of the model that are not affected by the other classes.

If, however, we apply \emph{both} clipping and noise with $S=1,
\sigma=0.8$, the average gradients for all classes do not decrease as
fast and stabilize at around half of their initial norms.  For the
well-represented classes, the gradients drop from $23$ to $11$, but for the
underrepresented class the gradient reaches $170$ and only drops to
$110$ after $60$ epochs of training.  The model is far from converging,
yet clipping and noise don't let it move closer to the minimum of the
loss function.  Furthermore, the addition of noise whose magnitude
is similar to the update vector prevents the clipped gradients of the
underrepresented class from sufficiently updating the relevant parts
of the model.  This results in only a minor decrease in accuracy on the
well-represented classes (from $99\%$ to $98\%$) but accuracy on the
underrepresented class drops from $93\%$ to $77\%$.  Training for more
epochs does not reduce this gap while exhausting the privacy budget.

Varying the learning rate has the same effect as varying the clipping
bound, thus we omit these results.


\paragraphbe{Noise multiplier $z$.} 
This parameter enforces a ratio between the clipping bound $S$ and noise
$\sigma$: $\sigma=zS$.  The lowest value of $z$ with the other parameters
fixed that still produces $\epsilon$ below $10$ is $z=0.7$. As discussed
above, the underrepresented class will have the gradient norm of $1$
and thus will be significantly impacted by such a large noise.

Figure~\ref{fig:mnist_experiments}(a) shows the accuracy of the model
under different $\epsilon$.  We experiment with different values of $S$
and $\sigma$ that result in the same privacy loss and report only the
best result.  For example, large values of $z$ require smaller $S$,
otherwise the model is destroyed by noise, but smaller $z$ lets us
increase $S$ and obtain a more accurate model.  In all cases, the
accuracy gap between the underrepresented and well-represented classes
is at least $20\%$ for the DP model vs.\ under $3\%$ for the non-DP model.

\paragraphbe{Batch size $b$.} 
Larger batches mitigate the impact of noise; also, prior
work~\cite{mcmahan2018general} recommends large batch sizes to help tune
performance of the model.  Figure~\ref{fig:mnist_experiments}(b) shows
that increasing the batch size decreases the accuracy gap at the cost
of increasing the privacy loss $\epsilon$.  Overall accuracy still drops.

\paragraphbe{Number of epochs $T$.} 
Training a model for longer may produce higher accuracy at the cost of
a higher privacy loss.  Figure~\ref{fig:mnist_experiments}(c) shows,
however, that longer training can still saturate the accuracy of the DP
model without matching the accuracy of the non-DP model.  Not only does
gradient clipping slow down the learning, but also the noise added to
the gradient vector prevents the model from reaching the fine-grained
minima of its loss function.  Similarly, in the iNaturalist model that
has many more parameters, added gradient noise degrades the model's
accuracy on the small classes.





\paragraphbe{Size of the underrepresented class.} 
In all preceding MNIST experiments, we unbalanced the classes with a
$12:1$ ratio, i.e., we used $500$ images of class ``8'' vs.\ $6,000$
images for the other classes.  Figure~\ref{fig:mnist_experiments}(d)
demonstrates that accuracy depends on the size of the underrepresented
group for both DP and non-DP models.  This effect becomes significant when
there are only $50$ images of the underrepresented class.  Clipping and
noise prevent the model from learning this class with $\epsilon<10$.

\section{Conclusion}

Gradient clipping and random noise addition, the core techniques at the
heart of differentially private deep learning, disproportionately affect
underrepresented and complex classes and subgroups.  As a consequence,
differentially private SGD has disparate impact: the accuracy of a model
trained using DP-SGD tends to decrease more on these classes and subgroups
vs.\ the original, non-private model.  If the original model is ``unfair''
in the sense that its accuracy is not the same across all subgroups,
DP-SGD exacerbates this unfairness.  We demonstrated this effect for
several image-classification and natural-language tasks and hope that
our results motivate further research on combining fairness and privacy
in practical deep learning models.

\paragraphbe{Acknowledgments.}
Many thanks to Omid Poursaeed for the iNaturalist experiments.  This
research was supported in part by the NSF grants 1611770, 1704296,
1700832, and 1642120, the generosity of Eric and Wendy Schmidt by
recommendation of the Schmidt Futures program, and a Google Faculty
Research Award.

\bibliographystyle{abbrv}
\bibliography{main}

\begin{thebibliography}{10}

\bibitem{abadi2016deep}
M.~Abadi, A.~Chu, I.~Goodfellow, H.~B. McMahan, I.~Mironov, K.~Talwar, and
  L.~Zhang.
\newblock Deep learning with differential privacy.
\newblock In {\em CCS}, 2016.

\bibitem{agarwal2018reductions}
A.~Agarwal, A.~Beygelzimer, M.~Dud{\'\i}k, J.~Langford, and H.~Wallach.
\newblock A reductions approach to fair classification.
\newblock In {\em ICML}, 2018.

\bibitem{beutel2017data}
A.~Beutel, J.~Chen, Z.~Zhao, and E.~H. Chi.
\newblock Data decisions and theoretical implications when adversarially
  learning fair representations.
\newblock In {\em {FAT/ML}}, 2017.

\bibitem{blodgett2016demographic}
S.~L. Blodgett, L.~Green, and B.~O'Connor.
\newblock Demographic dialectal variation in social media: A case study of
  {African-American English}.
\newblock In {\em EMNLP}, 2016.

\bibitem{blodgett2018twitter}
S.~L. Blodgett, J.~Wei, and B.~O’Connor.
\newblock {Twitter} universal dependency parsing for {African-American} and
  mainstream {American English}.
\newblock In {\em {ACL}}, 2018.

\bibitem{systematic_cnn}
M.~Buda, A.~Maki, and M.~A. Mazurowski.
\newblock A systematic study of the class imbalance problem in convolutional
  neural networks.
\newblock {\em Neural Networks}, 106:249--259, 2018.

\bibitem{buolamwini2018gender}
J.~Buolamwini and T.~Gebru.
\newblock Gender shades: Intersectional accuracy disparities in commercial
  gender classification.
\newblock In {\em FAT$\star$}, 2018.

\bibitem{chawla2002smote}
N.~V. Chawla, K.~W. Bowyer, L.~O. Hall, and W.~P. Kegelmeyer.
\newblock {SMOTE}: Synthetic minority over-sampling technique.
\newblock {\em JAIR}, 16:321--357, 2002.

\bibitem{Cui2019ClassBalancedLB}
Y.~Cui, M.~Jia, T.-Y. Lin, Y.~Song, and S.~J. Belongie.
\newblock Class-balanced loss based on effective number of samples.
\newblock In {\em CVPR}, 2019.

\bibitem{cummings2019compatibility}
R.~Cummings, V.~Gupta, D.~Kimpara, and J.~Morgenstern.
\newblock On the compatibility of privacy and fairness.
\newblock In {\em {FairUMAP}}, 2019.

\bibitem{douzas2018effective}
G.~Douzas and F.~Bacao.
\newblock Effective data generation for imbalanced learning using conditional
  generative adversarial networks.
\newblock {\em Expert Systems with Applications}, 91:464--471, 2018.

\bibitem{dwork2011differential}
C.~Dwork.
\newblock Differential privacy.
\newblock In {\em Encyclopedia of Cryptography and Security}, pages 338--340.
  Springer, 2011.

\bibitem{dwork2011firm}
C.~Dwork.
\newblock A firm foundation for private data analysis.
\newblock {\em CACM}, 54(1):86--95, 2011.

\bibitem{dwork2006calibrating}
C.~Dwork, F.~McSherry, K.~Nissim, and A.~Smith.
\newblock Calibrating noise to sensitivity in private data analysis.
\newblock In {\em TCC}, 2006.

\bibitem{elazar2018}
Y.~Elazar and Y.~Goldberg.
\newblock Adversarial removal of demographic attributes from text data.
\newblock In {\em EMNLP}, 2018.

\bibitem{geyer2017differentially}
R.~C. Geyer, T.~Klein, and M.~Nabi.
\newblock Differentially private federated learning: A client level
  perspective.
\newblock In {\em NeurIPS}, 2018.

\bibitem{hardt2016equality}
M.~Hardt, E.~Price, and N.~Srebro.
\newblock Equality of opportunity in supervised learning.
\newblock In {\em NIPS}, 2016.

\bibitem{he2016deep}
K.~He, X.~Zhang, S.~Ren, and J.~Sun.
\newblock Deep residual learning for image recognition.
\newblock In {\em CVPR}, 2016.

\bibitem{Horn_2018}
G.~V. Horn, O.~M. Aodha, Y.~Song, Y.~Cui, C.~Sun, A.~Shepard, H.~Adam,
  P.~Perona, and S.~Belongie.
\newblock The {iNaturalist} species classification and detection dataset.
\newblock In {\em CVPR}, 2018.

\bibitem{jagielski2018differentially}
M.~Jagielski, M.~Kearns, J.~Mao, A.~Oprea, A.~Roth, S.~Sharifi-Malvajerdi, and
  J.~Ullman.
\newblock Differentially private fair learning.
\newblock In {\em ICML}, 2019.

\bibitem{pmlr-v80-kearns18a}
M.~Kearns, S.~Neel, A.~Roth, and Z.~S. Wu.
\newblock Preventing fairness gerrymandering: Auditing and learning for
  subgroup fairness.
\newblock In {\em ICML}, 2018.

\bibitem{kearns1994toward}
M.~J. Kearns, R.~E. Schapire, and L.~M. Sellie.
\newblock Toward efficient agnostic learning.
\newblock {\em Machine Learning}, 17(2-3):115--141, 1994.

\bibitem{kuppam2019fair}
S.~Kuppam, R.~Mckenna, D.~Pujol, M.~Hay, A.~Machanavajjhala, and G.~Miklau.
\newblock Fair decision making using privacy-protected data.
\newblock {\em arXiv preprint arXiv:1905.12744}, 2019.

\bibitem{mcmahan2018general}
H.~B. McMahan, G.~Andrew, {\'{U}}.~Erlingsson, S.~Chien, I.~Mironov,
  N.~Papernot, and P.~Kairouz.
\newblock A general approach to adding differential privacy to iterative
  training procedures.
\newblock {\em arXiv:1812.06210}, 2018.

\bibitem{fedlearn_1}
H.~B. McMahan, E.~Moore, D.~Ramage, S.~Hampson, and B.~{Ag{\"{u}}era y Arcas}.
\newblock Communication-efficient learning of deep networks from decentralized
  data.
\newblock In {\em AISTATS}, 2017.

\bibitem{fedlearn_dp}
H.~B. McMahan, D.~Ramage, K.~Talwar, and L.~Zhang.
\newblock Learning differentially private recurrent language models.
\newblock In {\em ICLR}, 2018.

\bibitem{merler2019diversity}
M.~Merler, N.~Ratha, R.~S. Feris, and J.~R. Smith.
\newblock Diversity in faces.
\newblock {\em arXiv:1901.10436}, 2019.

\bibitem{mironov2017renyi}
I.~Mironov.
\newblock R{\'e}nyi differential privacy.
\newblock In {\em CSF}, 2017.

\bibitem{mohri2019agnostic}
M.~Mohri, G.~Sivek, and A.~T. Suresh.
\newblock Agnostic federated learning.
\newblock In {\em ICML}, 2019.

\bibitem{neelakantan2015adding}
A.~Neelakantan, L.~Vilnis, Q.~V. Le, I.~Sutskever, L.~Kaiser, K.~Kurach, and
  J.~Martens.
\newblock Adding gradient noise improves learning for very deep networks.
\newblock {\em arXiv:1511.06807}, 2015.

\bibitem{pascanu2012understanding}
R.~Pascanu, T.~Mikolov, and Y.~Bengio.
\newblock On the difficulty of training recurrent neural networks.
\newblock In {\em ICML}, 2013.

\bibitem{pytorch_link}
A.~Paszke, S.~Gross, S.~Chintala, G.~Chanan, E.~Yang, Z.~DeVito, Z.~Lin,
  A.~Desmaison, L.~Antiga, and A.~Lerer.
\newblock Automatic differentiation in {PyTorch}.
\newblock In {\em NIPS Workshops}, 2017.

\bibitem{pennington2014glove}
J.~Pennington, R.~Socher, and C.~Manning.
\newblock {GloVe}: Global vectors for word representation.
\newblock In {\em EMNLP}, 2014.

\bibitem{pytorchwordmodel}
\url{https://github.com/pytorch/}, 2019.
\newblock [Online; accessed 14-May-2019].

\bibitem{szegedy2015going}
C.~Szegedy, W.~Liu, Y.~Jia, P.~Sermanet, S.~Reed, D.~Anguelov, D.~Erhan,
  V.~Vanhoucke, and A.~Rabinovich.
\newblock Going deeper with convolutions.
\newblock In {\em CVPR}, 2015.

\bibitem{tf-privacy}
\url{https://github.com/tensorflow/privacy}, 2019.
\newblock [Online; accessed 14-May-2019].

\bibitem{yaghini2019disparate}
M.~Yaghini, B.~Kulynych, and C.~Troncoso.
\newblock Disparate vulnerability: On the unfairness of privacy attacks against
  machine learning.
\newblock {\em arXiv preprint arXiv:1906.00389}, 2019.

\bibitem{yeom2018privacy}
S.~Yeom, I.~Giacomelli, M.~Fredrikson, and S.~Jha.
\newblock Privacy risk in machine learning: Analyzing the connection to
  overfitting.
\newblock In {\em {CSF}}, 2018.

\bibitem{utkdataset}
Z.~Zhang, Y.~Song, and H.~Qi.
\newblock Age progression/regression by conditional adversarial autoencoder.
\newblock In {\em CVPR}, 2017.

\bibitem{zhu2019federated}
W.~Zhu, P.~Kairouz, H.~Sun, B.~McMahan, and W.~Li.
\newblock Federated heavy hitters discovery with differential privacy.
\newblock {\em arXiv:1902.08534}, 2019.

\end{thebibliography}

\end{document}